\pdfoutput=1

\documentclass{article}

\usepackage[most]{tcolorbox}
\tcbuselibrary{listings,breakable}
\makeatletter
\@ifundefined{tcb@tagsupport@enabled}{}{\def\tcb@tagsupport@enabled{false}}
\makeatother

\usepackage[preprint]{wenxiaobai}

\usepackage{times}
\usepackage{latexsym}
\usepackage{amsmath}
\usepackage{multirow}
\usepackage{booktabs}
\usepackage{makecell}
\usepackage{tabularx}
\usepackage{array}
\usepackage[T1]{fontenc}
\usepackage[utf8]{inputenc}
\usepackage{microtype}
\usepackage{inconsolata}
\usepackage{xcolor}
\usepackage{listings}
\usepackage{graphicx}
\usepackage{wrapfig}
\usepackage{hyperref}
\usepackage{multicol}
\usepackage{caption}
\usepackage{url}

\graphicspath{{figs/}}

\title{FS-Researcher: Test-Time Scaling for Long-Horizon Research Tasks with File-System-Based Agents}

\author{%
{\bf Chiwei Zhu\textsuperscript{\rm 1,2\S}, Benfeng Xu\textsuperscript{\rm 1,2\textdagger}, Mingxuan Du\textsuperscript{\rm 1}, Shaohan Wang\textsuperscript{\rm 1}} \\
{\bf Xiaorui Wang\textsuperscript{\rm 2}, Zhendong Mao\textsuperscript{\rm 1}, Yongdong Zhang\textsuperscript{\rm 1}} \\
\textsuperscript{1}University of Science and Technology of China \\ 
\textsuperscript{2}Metastone Technology \\ 
\texttt{\{tanz, benfeng\}@mail.ustc.edu.cn}}

\newcounter{savedfootnote}

\begin{document}
\maketitle

% --- Author footnotes (SynthQuestions-style symbols) ---
% Keep this local so it won't affect footnotes in the main body.
\begingroup
\setcounter{savedfootnote}{\value{footnote}}
\renewcommand{\thefootnote}{\fnsymbol{footnote}}
% 2 -> \dagger, 4 -> \S under \fnsymbol
\footnotetext[4]{Work done during the internship in Metastone Technology.}
\footnotetext[2]{Corresponding author. Project Lead.}
\endgroup
\setcounter{footnote}{\value{savedfootnote}}

\begin{abstract}
\hspace{2em}Deep research is emerging as a representative long-horizon task for large language model (LLM) agents. However, long trajectories in deep research often exceed model context limits, compressing token budgets for both evidence collection and report writing, and preventing effective test-time scaling. 
We introduce \textbf{FS-Researcher}, a file-system-based, dual-agent framework that scales deep research beyond the context window via a persistent workspace. Specifically, a \textbf{Context Builder} agent acts as a librarian which browses the internet, writes structured notes, and archives raw sources into a hierarchical knowledge base that can grow far beyond context length. A \textbf{Report Writer} agent then composes the final report section by section, treating the knowledge base as the source of facts. 
In this framework, the file system serves as a durable external memory and a shared coordination medium across agents and sessions, enabling iterative refinement beyond the context window.
Experiments on two open-ended benchmarks (DeepResearch Bench and DeepConsult) show that FS-Researcher achieves state-of-the-art report quality across different backbone models. Further analyses demonstrate a positive correlation between final report quality and the computation allocated to the Context Builder, validating effective test-time scaling under the file-system paradigm. The code and data are anonymously open-sourced at \url{https://github.com/Ignoramus0817/FS-Researcher}.
\end{abstract}

% Content placeholder - will be added in next sections
\section{Introduction}

Deep Research has recently emerged as a frontier and representative task for autonomous large language model (LLM) agents, demanding PhD-level expertise~\citep{OpenAIDeepResearch, GeminiDeepResearch}. Given an open-ended research query, deep research requires an agent to systematically collect evidence from the internet and synthesize it into a comprehensive report, which often involves navigating through hundreds of webpages and producing long reports containing more than 10K tokens.

The complexity of deep research poses a major challenge to agent design: models' context lengths are inherently limited, while long-horizon research tasks can easily exceed these capacities, halting further agent execution. This limitation prevents agents from allocating sufficient computation to tasks---the token budgets available for both information gathering and report writing are severely compressed, often falling short of what the task actually demands. As a result, static pipelines or single-agent workflows often suffer from incomplete coverage of relevant sources and produce lower-quality reports~\citep{storm, HFDeepResearch, webshaper, webthinker, deepresearcher}. 

To address this challenge, recent works typically reduce token consumption by offloading web browsing to sub-agents or summarizing tool observations, retaining only distilled key facts in the main agent context~\citep{TavilyDeepResearch, OpenDeepResearch, li2025webweaverstructuringwebscaleevidence, lei2025rhinoinsightimprovingdeepresearch, prabhakar2025enterprisedeepresearchsteerable}. While these methods extend the working trajectories of agents, they are temporary fixes that remain constrained by the hard limit of model context length. Moreover, in these approaches, internal states such as thoughts and tool observations are ephemeral consumables that are discarded once the agent loop terminates, hindering further scaling through iterative refinement.

\begin{wrapfigure}{r}{0.5\textwidth}
  \centering
  \includegraphics[width=0.9\linewidth]{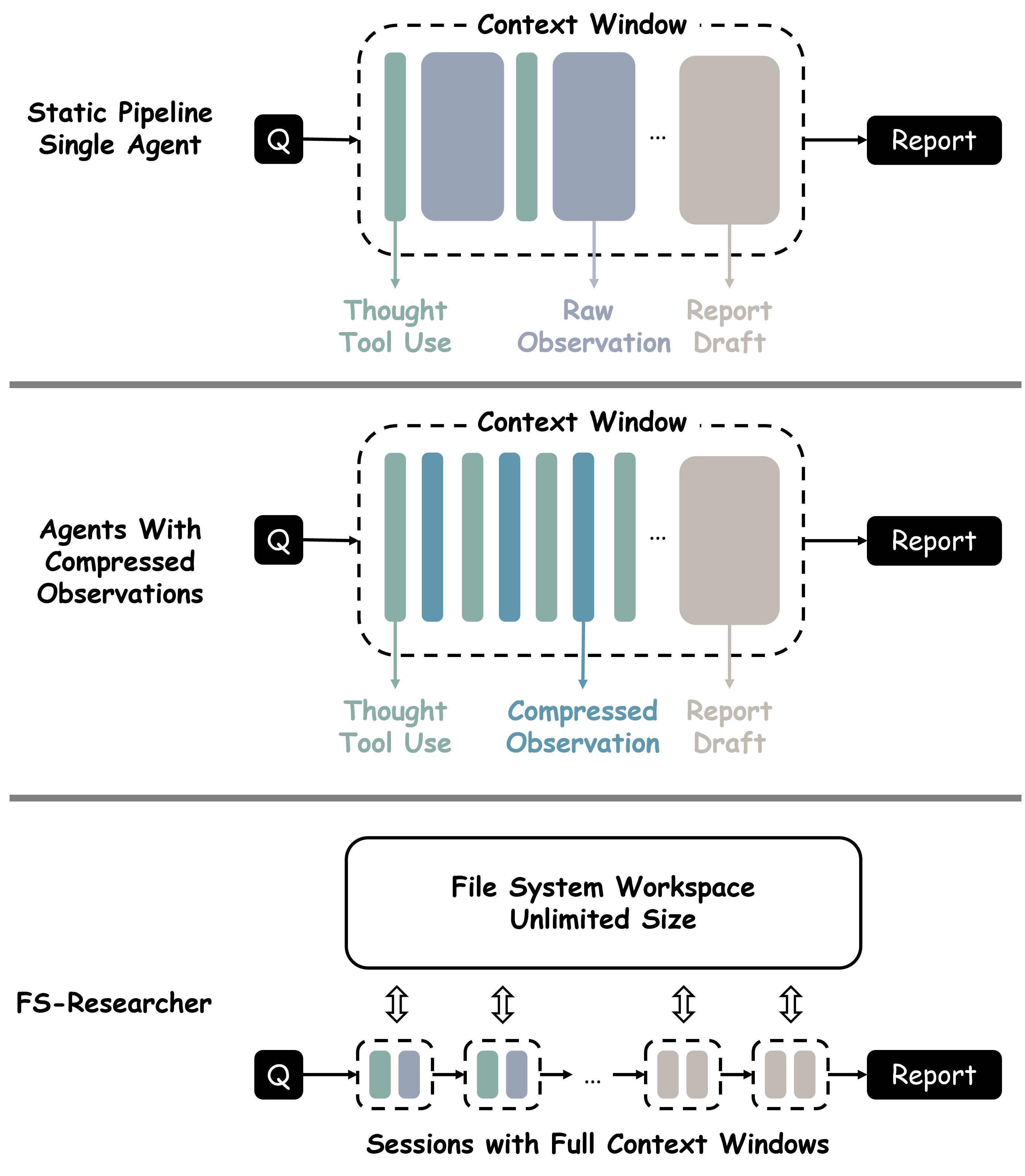}
  \caption{Different deep research paradigms: (1) \textbf{Top}: Static pipelines and naive single agents that put raw observations in the context; (2) \textbf{Middle}: Agents whose trajectories are extended by compressing the observations, while still bounded by the hard context limit; (3) \textbf{Bottom}: FS-Researcher, an agent framework built on top of an external file system workspace with unlimited context size.}
  \label{fig:comparison_paradigm}
\end{wrapfigure}

\begin{figure}[t]
  \centering
  \includegraphics[width=0.85\linewidth]{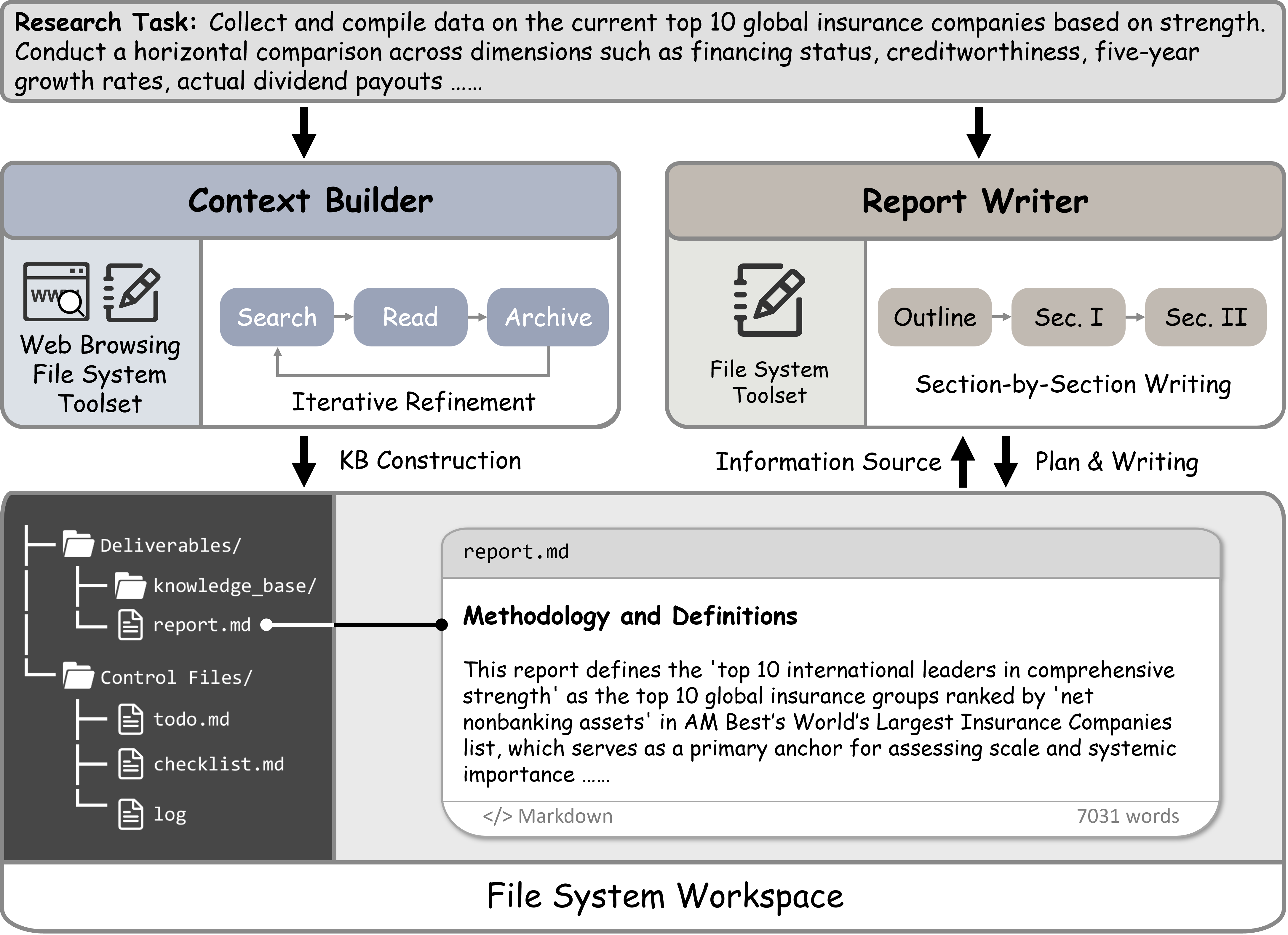}
  \caption{The framework of FS-Researcher.}
  \label{fig:framework}
\end{figure}

Recent progress in coding agents and AI-powered IDEs suggests that a file-system workspace is an effective substrate for long-horizon tool use and iterative development~\citep{yang2024sweagentagentcomputerinterfacesenable, Cursor, Claude-code}. Inspired by this paradigm yet addressing the unique nature of deep research—where agents must navigate noisy web content, extract and organize factual evidence, and synthesize coherent narratives—we propose \textbf{FS-Researcher}, a dual-agent framework that separates evidence accumulation from report composition. The first agent, \textbf{Context Builder}, functions as a librarian that browses the internet, reads potentially relevant documents, takes notes, and archives them into a hierarchically organized knowledge base whose size can far exceed context limits. The second agent, \textbf{Report Writer}, composes the report section by section, treating the knowledge base as its sole source of facts and loading relevant information on demand. Figure~\ref{fig:framework} shows the overview of FS-Researcher.

Our file-system-based workspace offers three key advantages: (1) it mirrors the native environment humans use for complex, long-horizon tasks, providing well-established interfaces for deep research; (2) it can store information far exceeding the model's context window, allowing on-demand access without context overflow; and (3) it makes intermediate artifacts (e.g., plans, error logs) persistent and revisitable, enabling iterative refinement across multiple agent sessions\footnote{Here we define \textbf{session} as a complete agent run from input prompt to final response.}. Notably, file I/O introduces negligible latency ($<$0.03\% of total wall-clock time; see Appendix~\ref{app:latency_profiling}). A comparison of file-system-based agents and existing paradigms is shown in Figure~\ref{fig:comparison_paradigm}.

As a result, our framework natively enables iterative refinement through the persistent workspace, thus allowing better token utilization for both information gathering and report writing. Extensive experiments demonstrate that FS-Researcher achieves state-of-the-art performance on open-ended deep research benchmarks across various backbone models. Further ablation studies reveal a positive correlation between the quality of the final report and the computation allocated to the \textbf{Context Builder} agent, indicating effective test-time scaling with the file-system paradigm.

Collectively, the contributions of this work are as follows:
\begin{itemize}
  \item We propose FS-Researcher, a dual-agent, file-system-based framework for solving long-horizon research tasks.
  \item We validate the effectiveness of our framework through extensive experiments.
  \item We demonstrate a positive correlation between deep research performance and the computation invested in context building.
\end{itemize}

\section{FS-Researcher}
FS-Researcher is a dual-agent framework that solves research tasks using a file-system-based workspace. The two agents also represent two stages: given a research topic, the \textbf{Context Builder} agent builds a comprehensive knowledge base, and then the \textbf{Report Writer} agent composes the report section by section. The agents share the same workspace and can refine the deliverables independently and iteratively.

\subsection{Architecture}
Before diving into the two agents, we introduce the common architecture that drives the whole framework, including three parts: tools, workflow, and workspace.

\paragraph{Tools.} FS-Researcher uses two types of tools: \textbf{file system} tools and \textbf{web browsing} tools, listed in Table~\ref{tab:tools}. We use Google SERP API and Jina AI API for \texttt{search\_web} and \texttt{read\_webpage} respectively.

\paragraph{Workflow.} FS-Researcher adopts a standard ReAct architecture for each agent, which can be formulated as follows:
\begin{align}
\label{eq:react}
T_i, A_i &= M_{\theta}(T_{j<i}, A_{j<i}, O_{j<i}, P) \\
O_i &= Execute(A_i)
\end{align}
\(T_i, A_i, O_i\) are the thought, action, and observation at the \(i\)-th step, respectively. \(M_{\theta}\) is the model with parameters \(\theta\). \(P\) is the prompt (system prompt and user query). \(Execute(A_i)\) is the tool implementation that executes the action \(A_i\) and returns the observation \(O_i\).

\paragraph{Workspace.} The workspace of FS-Researcher contains two types of files: \textbf{deliverables} and \textbf{control files}. All the files in the workspace are stored in Markdown format.
Deliverables are the final output files, which vary in form and convention depending on the task type. Detailed deliverables of each agent will be introduced in Section~\ref{sec:context-builder} and Section~\ref{sec:report-writer}.
Control files help the agents track the progress, and contain:
\begin{itemize}
  \item \textbf{Todos}: A list of tasks to be completed, each with a status of \texttt{[PENDING]}, \texttt{[IN-PROGRESS]}, or \texttt{[COMPLETE]}.
  \item \textbf{Checklist}: The acceptance criteria for a task, including file format rules, quality checks, etc.
  \item \textbf{Logs}: A log of the execution trajectory.
\end{itemize}
The framework natively supports multi-session workflow. 
At the beginning of each session, the agent inspects the current workspace, formulates a plan, and commences execution. 
During execution, the agent dynamically updates the todo file by modifying item statuses and adding, removing, or reordering tasks as needed.
Upon session completion, the agent evaluates the workspace against the checklist, re-marking any non-compliant items as \texttt{[IN-PROGRESS]}, and determines whether the overall task is complete. 
All inspection results, review findings, and session plans are recorded in the log file, which remains accessible to subsequent sessions and human collaborators, thereby facilitating iterative refinement. 
In our design, we let the agent generate todos autonomously and manually curate a static checklist. The checklists and a log example are shown in Appendix~\ref{app:checklists}, demonstrating how control files help recording the status and identifying issues.

\begin{table}[t]
  \centering
  \caption{Tools used in FS-Researcher.}
  \label{tab:tools}
  \resizebox{\linewidth}{!}{
    \begin{tabular}{c|c|l}
      \Xhline{1.0pt}
      \textbf{Type} & \textbf{Tool Name} & \multicolumn{1}{c}{\textbf{Description}} \\
      \hline
      \multirow{4}{*}{File System} & \texttt{ls} & List the files and sub-directories in target directory. \\
      \cline{2-3}
      & \texttt{grep} & A simplified version of UNIX grep command, search with a regular expression. \\
      \cline{2-3}
      & \texttt{read\_file} & Read a file. Pagination is supported with page size and page index as arguments. \\
      \cline{2-3}
      & \texttt{insert/delete/replace} & Modify certain lines in a file. Insert will write after the specified line by default. \\
      \Xhline{1.0pt}
      \multirow{2}{*}{Web Browsing} & \texttt{search\_web} & Search a query and return relevant URLs and summaries. \\
      \cline{2-3}
      & \texttt{read\_webpage} & Read a URL. Pagination is supported with page size and page index as arguments. \\
      \Xhline{1.0pt}
    \end{tabular}
  }
\end{table}

\subsection{Context Builder}
\label{sec:context-builder}
Given a research topic, the Context Builder works as a digital librarian that meticulously collects, distills, and archives information into a knowledge base (KB).

\begin{wrapfigure}{r}{0.42\linewidth}
\centering
\vspace{-10pt}
\begin{tcblisting}{
  colback=black!1,
  colframe=black!75,
  boxrule=0.6pt,
  arc=2pt,
  left=4pt,right=4pt,top=2pt,bottom=2pt,
  title={Knowledge Base Example},
  fonttitle=\bfseries\scriptsize,
  listing only,
  listing options={
    basicstyle=\ttfamily\tiny,
    columns=fullflexible,
    keepspaces=true,
    breaklines=true,
    aboveskip=0pt,
    belowskip=0pt
  }
}
./knowledge_base/
|-- sources/
|-- global_insurance_landscape/
|   |-- top10_rankings/
|   |   `-- leading_insurers_compilation.md
|   `-- metrics_definitions/
|       `-- strength_dimensions.md
|-- company_profiles/
|   |-- allianz/
|   |   |-- overview_and_financials_5y.md  
|   |   |-- dividends_and_payouts.md
|   |   |-- credit_ratings.md
|   |   `-- china_strategy_and_presence.md
|   |-- ...
|   |-- ...
`-- comparative_analysis/
    |-- financing_structure_comparison.md  
    |-- reputation_and_ratings_comparison.md
    |-- growth_5y_comparison.md
    |-- dividends_payouts_comparison.md  
    |-- china_potential_assessment.md
    `-- future_top_assets_candidates_2_3.md
\end{tcblisting}
\caption{Knowledge base example.}
\label{fig:kb-structure}
\vspace{-10pt}
\end{wrapfigure}

The deliverables of this agent include one file (\texttt{index.md}) and two directories (\texttt{knowledge\_base/} and \texttt{sources/}).
The \texttt{index.md} is like the ``Table of Content'' of the KB, which contains two parts: (1) the deconstruction of the research topic, and (2) the hierarchical structure of the KB. From the \texttt{index.md}, the agent or human collaborators can get an overview of what the KB is built for and how it is organized, and navigate to specific information sources more efficiently. \texttt{Todos} are created and updated along with the modification of \texttt{index.md}, which guides the browsing process.
The \texttt{knowledge\_base/} directory contains the notes written by the Context Builder when browsing the internet, organized in a tree-like structure. The names of folders and files are descriptive, reflecting the semantic relationships between the deconstructed topics. 
The \texttt{sources/} contains the raw webpages archived from the internet. For trackability, each statement in the notes of \texttt{knowledge\_base/} comes with a citation that points to a file in the \texttt{sources/} directory.

During the context building stage, the Context Builder browses the internet with \texttt{search\_web} and \texttt{read\_webpage} tools, updates the \texttt{index.md}, distills key information into notes in the \texttt{knowledge\_base/} directory, and archives raw webpages in the \texttt{sources/} directory. Note that this workflow is not linear, i.e. first deconstruct the topic, design a target structure of KB and fill the folders with files. Instead, the \texttt{index.md} and \texttt{knowledge\_base/} directory are dynamically updated as the agent browses the internet and gradually forms its understanding of the topic. 

At the end of each session, the Context Builder conducts a review against the checklist, identifying any potential errors, gaps, or conflicts in the knowledge base. If any issues are found, corresponding items are marked as \texttt{[IN-PROGRESS]} and recorded in the \texttt{log} file. The Context Builder can iteratively refine the knowledge base until it reaches the session budget limit (i.e., the maximum number of sessions that the Context Builder is allowed to run) or does not identify any issue in the review.
Figure~\ref{fig:kb-structure} shows an example of the knowledge base, full structure in Appendix~\ref{app:kb}.

\subsection{Report Writer}
\label{sec:report-writer}
Once the Context Builder marks the knowledge base as complete, the Report Writer takes over the workspace and starts to compose the report. In this stage, we remove the web browsing tools and let Report Writer treat the knowledge base built by Context Builder as the only source of facts. The deliverable of this stage is \texttt{report.md}.

A critical observation is that if the whole report is written in one-shot generation, it tends to read like a mere list of facts, lacking explanation and in-depth analysis. 
Therefore, we adopt a multi-session writing process, where the Report Writer creates an outline file in the first writing session, and chooses exactly one section to compose in subsequent sessions. The outline also serves as the \texttt{todo} file for the Report Writer, where each section carries a status of \texttt{[PENDING]}, \texttt{[IN-PROGRESS]}, or \texttt{[COMPLETE]}.
Upon the completion of a section, the Report Writer performs a section-level review against the section-level checklist. The status of the current section is changed to \texttt{[COMPLETE]} only when the self-check passes.
After all sections are completed, an overall review is conducted using the report-level checklist. If flaws are identified, the corresponding sections are marked as \texttt{[IN-PROGRESS]} again.

The Report Writer continuously executes until the entire report is finished and passes all reviews. There is no budget limit in this stage.

\begin{table*}[t]
  \centering
  \caption{Performance on DeepResearch Bench. \texttt{Comp.}, \texttt{Instr.}, \texttt{Eff.c.} and \texttt{C.acc.} denotes comprehensiveness, instruction following, effective citations and citation accuracy respectively. The best performance is highlighted in bold and the second best is underlined.}
  \label{tab:deepresearchbench}
  \resizebox{\linewidth}{!}{
    \begin{tabular}{llccccccc}
    \toprule
    \multirow{2}{*}{\textbf{Category}} & \multirow{2}{*}{\textbf{Method}} & \multicolumn{5}{c}{\textbf{RACE}} & \multicolumn{2}{c}{\textbf{FACT}} \\
    \cmidrule(lr){3-7}\cmidrule(lr){8-9}
    &  & \textbf{Comp.} & \textbf{Insight} & \textbf{Instr.} & \textbf{Readability} & \textbf{Overall} & \textbf{Eff. c.} & \textbf{C.acc.} \\
    \midrule
    \multirow{3}{*}{Proprietary} 
    & Claude-DeepResearch                     & 45.34 & 42.79 & 47.58 & 44.66 & 45.00 & - & - \\
    & OpenAI-DeepResearch                     & 46.46 & 43.73 & 49.39 & 47.22 & 46.45 & 39.79 & 75.01 \\
    & Gemini-2.5-Pro-DeepResearch             & 49.51 & 49.45 & 50.12 & 50.00 & 49.71 & \textbf{165.34} & \textbf{78.30} \\
    \midrule
    \multirow{4}{*}{Open Source} 
    & LangChain-Open-Deep-Research (GPT-5)            & 50.06 & 50.76 & 51.31 & 49.72 & 50.60 & 22.44 & 34.74 \\
    & EnterpriseDeepResearch (Gemini-2.5-Pro)         & 49.70 & 51.24 & 50.52 & 50.61 & 50.62 & - & 72.50 \\
    & WebWeaver (Qwen3-235B-A22B-Instruct-2507)       & 51.45 & 51.39 & 50.26 & 48.98 & 50.80 & \underline{152.70} & 75.72 \\
    & RhinoInsight (Gemini-2.5-Pro)                   & 50.51 & 51.45 & 51.72 & 50.00 & 50.92 & - & - \\
    \midrule
    \multirow{3}{*}{Ours} & FS-Researcher (Gemini-2.5-Pro)    & 51.25 & \underline{55.03} & \underline{52.38} & 50.77 & 52.51 & 56.67 & \underline{78.20} \\
    & FS-Researcher (GPT-5)    & \underline{51.96} & 54.44 & 52.14 & \underline{51.26} & \underline{52.76} & 113.23 & 60.04 \\
    & FS-Researcher (Claude-Sonnet-4.5)    & \textbf{54.25} & \textbf{55.85} & \textbf{52.47} & \textbf{51.54} & \textbf{53.94} & 139.91 & 76.17 \\
    \bottomrule
    \end{tabular}
  }
\end{table*}

\section{Experiments}
In this section, we evaluate the performance of FS-Researcher on open-ended deep research benchmarks. Then we conduct detailed analyses on the scaling effect of the framework, the impact of different modules, and present a concrete showcase.

\subsection{Experimental Setups}
\paragraph{Benchmarks.} We evaluate the performance of FS-Researcher on two widely-used deep research benchmarks: DeepResearch Bench~\citep{du2025deepresearchbenchcomprehensivebenchmark} and DeepConsult~\citep{DeepConsult}. DeepResearch Bench consists of 100 PhD-level research tasks across 22 distinct fields and evaluate the deep research systems based on the report quality and citation accuracy. DeepConsult contains 103 research queries mainly about business and consulting. Following \texttt{Gemini-DeepResearch}, DeepConsult evaluates reports based on 4 dimensions: instruction following, comprehensiveness, completeness and writing quality. Both benchmarks use LLM-as-a-judge evaluation. To additionally validate our framework with verifiable metrics, we also evaluate on BrowseComp~\citep{browsecomp}, an agentic search benchmark comprising 1,266 complex but answer-verifiable questions that require extensive web browsing; we test on a randomly sampled subset of 100 queries. Details of the benchmarks are shown in Appendix~\ref{app:benchmark}. All the scores are the average of 3 test runs.

\paragraph{Baselines.} We compare the performance of FS-Researcher against two types of state-of-the-art deep research systems: (1) strong proprietary deep research products including OpenAI Deep Research~\citep{OpenAIDeepResearch}, Claude-Research~\citep{ClaudeDeepResearch}, Gemini-2.5-Pro-DeepResearch~\citep{GeminiDeepResearch} and (2) open-source deep research systems and recently released papers, including LangChain-Open-Deep-Research~\citep{OpenDeepResearch}, WebWeaver~\citep{li2025webweaverstructuringwebscaleevidence}, RhinoInsight~\citep{lei2025rhinoinsightimprovingdeepresearch} and EnterpriseDeepResearch~\citep{prabhakar2025enterprisedeepresearchsteerable}.

\subsection{Main Results}
Table~\ref{tab:deepresearchbench} shows the results on DeepResearch Bench, on which FS-Researcher with \texttt{Claude-Sonnet-4.5} reaches \textbf{53.94} RACE and significantly outperforms the strongest baseline (RhinoInsight, \textbf{+3.02}). Notably, the comprehensiveness and insight are improved by large margins (\textbf{+3.74}/\textbf{+4.4} over the previous best), which highlights the ability of FS-Researcher on broadly collecting evidences and synthesizing in-depth analyses. FS-Researcher \texttt{Claude-Sonnet-4.5} also shows competitive performance on citation accuracy, only left behind \texttt{Gemini-2.5-Pro-DeepResearch}.

Importantly, our gain is not solely attributable to a stronger backbone model. Under the same GPT-5 backbone, FS-Researcher improves RACE by \textbf{+2.16} over \texttt{LangChain Open Deep Research}; under the same Gemini-2.5-Pro backbone, it outperforms RhinoInsight by \textbf{+1.59} RACE. Furthermore, the agent harness itself contributes a substantial share of the gain: Gemini-2.5-Pro equipped with only a search tool scores 31.9 RACE, whereas the same model inside its official harness reaches 49.71 RACE, and FS-Researcher pushes this further to 52.51 RACE (see Appendix~\ref{app:harness_contribution} for the full breakdown). Together, these results confirm that the file-system-based two-stage workflow and persistent knowledge base provide a complementary, framework-level benefit beyond backbone choices. FS-Researcher with \texttt{GPT-5} does not achieve as high citation accuracy as \texttt{Claude-Sonnet-4.5} because \texttt{GPT-5} tends to stack several citations at the end of a paragraph, which might result in the misalignment between cited sources and extracted fact statements.

\begin{table*}[t]
  \centering
  \caption{Performance on DeepConsult. $^\dag$ Results for Claude are based on sampled 20 queries due to budget limit.}
  \label{tab:deepconsult}
  \small
  \begin{tabular}{llcccc}
  \toprule
  \textbf{Category} & \textbf{Method} & \textbf{Win(\%)} & \textbf{Tie(\%)} & \textbf{Lose(\%)} & \textbf{Avg. score} \\
  \midrule
  \multirow{3}{*}{Proprietary} 
  & Claude-DeepResearch                     & 25.0 & 38.89 & 36.11 & 4.6 \\
  & OpenAI-DeepResearch                     & 0.00 & 100.00 & 0.00 & 5.00 \\
  & Gemini-2.5-Pro-DeepResearch             & 61.27 & 31.13 & 7.60 & 6.70 \\
  \midrule
  \multirow{4}{*}{Open Source} 
  & EnterpriseDeepResearch (Gemini-2.5-Pro)         & 71.57 & 19.12 & \underline{9.31} & 6.82 \\
  & RhinoInsight (Gemini-2.5-Pro)                   & 68.51 & \underline{11.02} & 20.47 & 6.82 \\
  & WebWeaver (Qwen3-235B-A22B-Instruct-2507)       & 66.16 & 12.14 & 21.68 & 6.94 \\
  \midrule
  \multirow{3}{*}{Ours} & FS-Researcher (Gemini-2.5-Pro)     & 74.88 & 16.01 & \underline{9.11} & 7.62 \\
   & FS-Researcher (GPT-5)     & \underline{73.28} & 18.97 & \textbf{7.76} & \underline{7.26} \\
   & FS-Researcher (Claude-Sonnet-4.5)$^\dag$       & \textbf{80.00} & \textbf{10.42} & 9.58 & \textbf{8.33} \\
  \bottomrule
  \end{tabular}
\end{table*}

Table~\ref{tab:deepconsult} shows the results on DeepConsult, displaying similar trends with DeepResearch Bench. FS-Researcher with \texttt{Claude-Sonnet-4.5} attains the highest win rate (\textbf{80.00}\%) and the best average score (\textbf{8.33}), while substantially reducing losses (\textbf{9.58}\%). Detailed results for each dimension are listed in Appendix~\ref{app:deepconsult}.

To complement the LLM-as-a-judge evaluations above, we also evaluate on \textbf{BrowseComp}~\citep{browsecomp}, an answer-verifiable agentic search benchmark. As shown in Table~\ref{tab:browsecomp}, FS-Researcher outperforms the corresponding official agent harness under both backbones, demonstrating that the framework's advantage extends beyond LLM-as-judge settings to objectively verifiable metrics.

\begin{table}[t]
  \centering
  \caption{BrowseComp accuracy on a random 100-query subset. FS-Researcher outperforms official agent harnesses with both backbones.}
  \small
  \begin{tabular}{lc}
  \toprule
  \textbf{Method} & \textbf{Accuracy (\%)} \\
  \midrule
  Claude-Sonnet-4.5 (official harness) & 43.9 \\
  FS-Researcher (Claude-Sonnet-4.5) & \textbf{55.0} \\
  \midrule
  GPT-5 (official harness) & 54.9 \\
  FS-Researcher (GPT-5) & \textbf{68.0} \\
  \bottomrule
  \end{tabular}
  \label{tab:browsecomp}
\end{table}

Together, these results across three benchmarks demonstrate that FS-Researcher consistently improves research quality in both open-ended report generation and answer-verifiable information seeking.

To better understand the dynamics of our two-stage workflow, we analyze the tool trajectories of the first three iterations in both stages, and visualize the frequency--position heatmaps. We observe patterns that align with our design (e.g., search-before-read, early workspace inspection, and late self-check editing); detailed analysis and figures are provided in Appendix~\ref{app:tool_usage_patterns}.

\section{Analyses}
In this section, we verify the scaling effect of the Context Builder in FS-Researcher and analyze the impact of different modules. We randomly sample 10 queries from the DeepResearch Bench test set and use \texttt{GPT-5} as the backbone for experiments.

\subsection{Scaling Effect of Context Builder}
\label{sec:scaling_effect}
We study whether FS-Researcher exhibits test-time scaling by varying the computation allocated to the \textbf{Context Builder}. Concretely, we run the Context Builder for different numbers of rounds (3/5/10) before invoking the Report Writer. We additionally report the cost and web-browsing tool usage under different rounds in Appendix~\ref{app:cost}, and show that context compression with a smaller summarizer model can reduce Context Builder cost by 47\% with negligible quality loss (Appendix~\ref{app:context_compression}).

% \begin{wrapfigure}{r}{0.4\linewidth}
%   \centering
%   \vspace{-10pt}
%   \includegraphics[width=\linewidth]{scaling.png}  
%   \caption{DeepResearch Bench scores of FS-Researcher with 3-10 rounds of context building.}
%   \label{fig:scaling_drb}
%   \vspace{-10pt}
% \end{wrapfigure}

\begin{figure}[t]
  \centering
  \vspace{-10pt}
  \includegraphics[width=\linewidth]{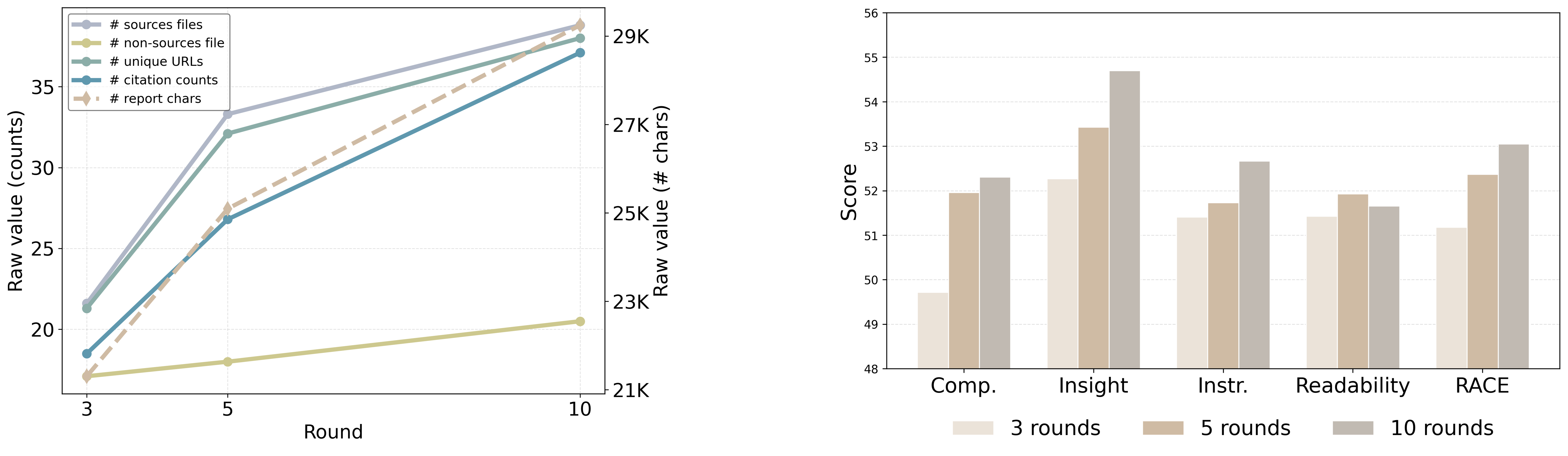}  
  \caption{\textbf{Left}: KB statistics under 3-10 rounds of context-building. The number of characters in report corresponds to the y-axis on the right. \textbf{Right}: DeepResearch Bench scores of FS-Researcher with 3-10 rounds of context building.}
  \label{fig:scaling}
  \vspace{-10pt}
\end{figure}

\paragraph{Knowledge Base.} Figure~\ref{fig:scaling} (left) summarizes how the KB grows with more context-building rounds. As the budget increases from 3 to 10 rounds, the archived information and notes keep increasing. Meanwhile, the downstream report becomes longer and contains more citations, indicating that richer KBs enable denser and better-grounded synthesis. Notably, KB growth shows diminishing returns: most gains occur from 3 to 5 rounds (e.g., +11.7 sources / +10.8 URLs) with smaller increments from 5 to 10 rounds (+5.5 / +5.9), suggesting that the KB becomes progressively more complete and the remaining information gaps shrink as more rounds are performed. The original aggregated statistics are provided in Appendix~\ref{app:scaling_data}.

\paragraph{Report.} Figure~\ref{fig:scaling} (right) reports the average performance over the 10 sampled queries. Increasing the number of rounds consistently improves all quality dimensions except for \texttt{Readability}, indicating that additional computation invested in building a higher-quality knowledge base translates into better final reports. The original scores are listed in Appendix~\ref{app:scaling_data}.
Interestingly, \texttt{Readability} peaks at 5 rounds (51.93) and slightly drops at 10 rounds (51.66). We note that comprehensiveness and readability are \emph{conceptually orthogonal}---the former measures \emph{what} information is covered, the latter measures \emph{how} it is communicated---and FS-Researcher with Claude-Sonnet-4.5 simultaneously achieves the highest scores on both (Table~\ref{tab:deepresearchbench}). The observed tradeoff is a confounding artifact: a larger KB leads the Report Writer to adopt a denser, more technical writing style, which incidentally harms readability. Because this is a presentation-level issue, a targeted post-hoc rephrasing pass can recover readability while preserving comprehensiveness (see Appendix~\ref{app:readability_rephrasing}).

\subsection{Module Ablations}
To study the effectiveness of several critical designs in FS-Researcher, we conduct 3 ablation experiments on \textbf{Persistent Workspace}, \textbf{Dual-Agent}, and \textbf{Section-wise Writing}. Table~\ref{tab:module_ablations} shows the result of the three ablation experiments.

\paragraph{Persistent Workspace.} We remove the workspace design that helps status sharing and iterative refinement from the FS-Researcher. Specifically, we remove the control files from both stages and replace the structured workspace in Context Builder with a flat note. In such setting, the agents can still iteratively work but only based on the final deliverables, without fine-grained perception of task progress. The budget of the two agents are the same in the main experiments, i.e. 3 rounds for Context Builder and unlimited rounds for Report Writer.
As shown in Table~\ref{tab:module_ablations}, removing the persistent workspace leads to a clear overall degradation (52.76 $\rightarrow$ 48.69 RACE). The drop is most pronounced on \texttt{Insight} (-7.95) and is accompanied by a consistent decline on \texttt{Comprehensiveness} (-3.58), indicating that without explicit task state and a structured knowledge base, the agents are less effective at identifying information gaps, consolidating evidence, and refining hypotheses across iterations. In comparison, \texttt{Instruction Following} and \texttt{Readability} decrease less, suggesting that the workspace primarily contributes to deeper reasoning and coverage rather than surface-level format compliance.

\paragraph{Dual-Agent.} We merge the Context Builder and Report Writer into a single agent, which browses the internet and writes the report in a same working session. For each session, the agent should provide ready-to-go deliverables. We run the single agent for 3 rounds.
As shown in Table~\ref{tab:module_ablations}, this ablation causes the largest overall degradation (52.76 $\rightarrow$ 42.41 RACE), with a particularly severe drop on \texttt{Insight} (-16.89) and \texttt{Comprehensiveness} (-11.06). This suggests that interleaving evidence acquisition and report drafting within the same session encourages premature synthesis and shallow exploration: the agent starts writing before the workspace is sufficiently grounded and struggles to maintain a stable, structured knowledge base while producing polished report. More fundamentally, the single-agent workflow must continuously split its effective context and generation capacity between browsing/notes and long-form writing, leaving insufficient room for either exhaustive evidence collection or careful section-level reasoning, which directly harms report quality.

\begin{table*}[!htbp]
  \centering
  \caption{The experimental results of the module ablations.}
  \label{tab:module_ablations}
  \small
  \begin{tabular}{lccccc}
  \toprule
  \textbf{Setting} & \textbf{Comp.} & \textbf{Insight} & \textbf{Instr.} & \textbf{Readability} & \textbf{RACE} \\
  \midrule
  FS-Researcher (GPT-5)    & 51.96 & 54.44 & 52.14 & 51.26 & 52.76 \\
   - Persistent Workspace & 48.38 (-3.58) & 46.49 (-7.95) & 50.78 (-1.36) & 49.92 (-1.34) & 48.69 (-4.07) \\
   - Dual-Agent & 40.90 (-11.06) & 37.55 (-16.89) & 46.30 (-5.84) & 44.78 (-6.48) & 42.41 (-10.35) \\
   - Section-wise Writing & 47.06 (-4.90) & 45.64 (-8.80) & 50.50 (-1.64) & 46.46 (-4.80) & 47.63 (-5.13) \\
  \bottomrule
  \end{tabular}
\end{table*}

\paragraph{Section-wise Writing.} We let the Report Writer write the whole report in one-shot generation, instead of writing section-by-section. We directly use the KB constructed in the main experiments and write on top of them.
The RACE scores show a sharp decline (52.76 $\rightarrow$ 47.63) in such setting. 
The degradation is consistent across all dimensions, with the largest drop on \texttt{Insight} (-8.80) and notable decreases on \texttt{Comprehensiveness} (-4.90) and \texttt{Readability} (-4.80), indicating that long-form one-shot writing often loses analytical depth and structural clarity. 
In contrast, section-wise writing provides frequent opportunities to re-ground on the outline and the constructed knowledge base, enabling local planning and self-correction that improves the overall report quality. Instruction Following drops the least, as the content of the report is also complete when written in one shot, addressing most of the research task.

\subsection{Case Study}
We present a showcase that clearly reflects the evolution of KBs and resulting reports under different context-building rounds (3/5/10 rounds). The research query is: \emph{``Gather information on the world's top 10 insurance companies by overall strength. Compare them across financing, reputation, 5-year growth, historical dividends, and future potential in the Chinese market; then select the 2--3 companies most likely to rank highest in future total assets.''} The different KBs and report excerpts are listed in Appendix~\ref{app:case_study}.

\paragraph{KB growth with more rounds.}
As the Context Builder allocated more rounds, the KB becomes larger and more structured:
From 5 to 10 rounds, while the number of archived sources increases only a few (54 $\rightarrow$ 59), the number of evidence notes increases considerably (75 $\rightarrow$ 98), including more cross-source comparisons and analyses.

\paragraph{How KB growth changes the report.}
Reports become progressively more evidence-grounded and modular: with 3 rounds, the report already identifies the top-10 set and a preliminary shortlist but has limited metric-by-metric synthesis; with 5 rounds, it adds explicit methodology and dimension-wise comparisons; with 10 rounds, it further consolidates trade-offs by grounding claims in reusable KB modules.

\paragraph{Readability.}
As information density increases, reports become more technical: domain terms (e.g., solvency frameworks and capital ratios) and inline citations become denser, which makes the report harder to follow.
Consistently, the readability score decreases monotonically (59.32 $\rightarrow$ 55.49 $\rightarrow$ 54.79), clearly supporting our assumption in Section~\ref{sec:scaling_effect}.

\section{Related Works}
Recent progress in models' reasoning and tool using capability has spawned Deep Research Agents that autonomously perform multi-step information gathering and report synthesis~\citep{openai2024openaio1card,deepseekai2025deepseekr1incentivizingreasoningcapability, lin2025understandingtoolintegratedreasoning, patil2025bfcl, yao2024taubenchbenchmarktoolagentuserinteraction, guo2025mcpagentbenchevaluatingrealworldlanguage}. While proprietary products have shown impressive, human-level performance as pioneers~\citep{OpenAIDeepResearch, ClaudeDeepResearch, GeminiDeepResearch}, the technique behind them remains largely opaque, leaving researchers with limited understanding of how to construct agents for long-horizon research tasks.

Open-source efforts attempt to bridge this gap by building systems with reproducible workflows. Early approaches typically rely on static pipelines or simple single-agent workflows~\citep{storm, HFDeepResearch, webshaper, webthinker, deepresearcher}, which often struggle to scale to extra long-horizon settings: as the trajectory grows, the limited context window forces thoughts, observations, and report draft to compete for tokens, which can lead to incomplete source coverage, premature synthesis, and brittle behavior. To mitigate context pressure, subsequent works compress the agent workflow context to extend the number of steps within a session, with a representative strategy being to summarize tool observations and keep only distilled key facts in the main agent context~\citep{TavilyDeepResearch, OpenDeepResearch, li2025webweaverstructuringwebscaleevidence, lei2025rhinoinsightimprovingdeepresearch, prabhakar2025enterprisedeepresearchsteerable}. While effective at prolonging trajectories, such compression introduces an inherently lossy bottleneck: fine-grained evidence and provenance may be dropped, errors can accumulate across summarization stages, and the agent remains bounded by the hard context limit. In contrast, our approach aims to enable effective test-time scaling for deep research by externalizing research state and evidence into a persistent, well-structured workspace that can be revisited and refined across sessions, allowing additional computation to be allocated.

\section{Conclusion}

In this paper, we presented \textbf{FS-Researcher}, a file-system-based dual-agent framework for long-horizon deep research tasks that scales beyond the context window via a persistent workspace. FS-Researcher separates research into two stages: a \textbf{Context Builder} that decomposes the query, collects evidence from the web, and curates a hierarchical, citation-grounded knowledge base; and a \textbf{Report Writer} that composes the final report section by section using on-demand retrieval from the workspace. Experiments on DeepResearch Bench and DeepConsult show that FS-Researcher achieves state-of-the-art report quality across different backbones, and our analyses further demonstrate a positive relationship between report quality and the computation allocated to context building, validating effective test-time scaling under the file-system paradigm.

\section*{Limitations}
The main limitation of this work lies in its dependence on relatively strong foundational models. Deep research is inherently a demanding setting: it requires robust multi-turn planning, broad web search, and long-form writing with coherent structure. In our framework, the file system operations further require strong reasoning and function calling abilities. As a result, smaller or less capable backbones (e.g., \texttt{gpt-5-mini}) may exhibit shorter trajectories and more frequent premature stopping, which in practice translates to requiring more sessions to reach comparable coverage. They may also be more prone to vulnerabilities in file operations (e.g., incorrect edits, inconsistent state updates, or erroneous tool use), reducing overall task success rates. That said, our preliminary experiments show that \texttt{GPT-5-mini}, when given additional context-building rounds, achieves performance comparable to OpenAI-DeepResearch at a substantially lower cost (\$2.51 vs.\ \$6.10/query), suggesting the framework remains usable beyond top-tier LLMs (Appendix~\ref{app:smaller_models}). Designing a less demanding framework that better supports smaller models is an important direction for future work.

\section*{Ethical Considerations}

FS-Researcher relies on web-sourced content; despite citation grounding, it may still propagate inaccurate, biased, or outdated information, which could mislead downstream decisions. Persisting retrieved materials and intermediate notes in a file-system workspace may inadvertently store sensitive or copyrighted content and, in untrusted environments, increase the attack surface for prompt injection or malicious pages that attempt to influence tool actions.

\section*{Acknowledgments}
This research is supported by Artificial Intelligence-National Science and Technology Major Project 2023ZD0121200, Science Fund for Creative Research Groups under Grant 62121002 and Fundamental and Interdisciplinary Disciplines Breakthrough Plan of the Ministry of Education of China under No.JYB2025XDXM103.

\newpage
\bibliography{custom}
\bibliographystyle{acl_natbib}

\newpage
\appendix

\section{Knowledge Base Sample}
\label{app:kb}
Below is a structured knowledge base constructed by Context Builder. The corresponding query is from DeepResearch Bench: ``Gather information on the world's top 10 insurance companies by overall strength. Compare them across the following metrics: financing, reputation, 5-year growth, historical dividends, and future potential in the Chinese market. Finally, assess and select the 2-3 companies most likely to rank highest in future total assets.''.

\begin{center}
\begin{minipage}{0.85\linewidth}
\begin{tcblisting}{
  colback=black!1,
  colframe=black!75,
  boxrule=0.6pt,
  arc=2pt,
  left=6pt,right=6pt,top=6pt,bottom=6pt,
  title={Knowledge Base Structure Example},
  fonttitle=\bfseries\small,
  listing only,
  listing options={
    basicstyle=\ttfamily\scriptsize,
    columns=fullflexible,
    keepspaces=true,
    breaklines=true
  }
}
./knowledge_base/
|-- sources/
|-- global_insurance_landscape/
|   |-- top10_rankings/
|   |   `-- leading_insurers_compilation.md
|   `-- metrics_definitions/
|       `-- strength_dimensions.md
|-- company_profiles/
|   |-- allianz/
|   |   |-- overview_and_financials_5y.md  
|   |   |-- dividends_and_payouts.md
|   |   |-- credit_ratings.md
|   |   `-- china_strategy_and_presence.md
|   |-- axa/
|   |   |-- overview_and_financials_5y.md
|   |   |-- dividends_and_payouts.md
|   |   `-- credit_ratings.md
|   |-- ping_an/
|   |   |-- overview_and_financials_5y.md
|   |   |-- dividends_and_payouts.md
|   |   `-- credit_ratings.md
|   |-- china_life/
|   |   |-- overview_and_financials_5y.md
|   |   |-- dividends_and_payouts.md
|   |   `-- credit_ratings.md
|   |-- aia/
|   |   |-- overview_and_financials_5y.md
|   |   |-- dividends_and_payouts.md
|   |   |-- china_expansion.md
|   |   `-- credit_ratings.md
|   |-- zurich/
|   |   |-- overview_and_financials_5y.md
|   |   |-- dividends_and_payouts.md
|   |   `-- credit_ratings.md
|   |-- generali/
|   |   |-- overview_and_financials_5y.md
|   |   |-- china_strategy_and_presence.md
|   |   `-- credit_ratings.md
|   |-- prudential_plc/
|   |   |-- overview_and_financials_5y.md
|   |   |-- dividends_and_payouts.md
|   |   `-- credit_ratings.md
|   |-- prudential_financial/
|   |   |-- overview_and_financials_5y.md
|   |   |-- dividends_and_payouts.md
|   |   `-- credit_ratings.md
|   |-- metlife/
|   |   |-- overview_and_financials_5y.md
|   |   |-- dividends_and_payouts.md
|   |   `-- credit_ratings.md
|   `-- manulife/
|       |-- overview_and_financials_5y.md
|       |-- dividends_and_payouts.md
|       `-- credit_ratings.md
`-- comparative_analysis/
    |-- financing_structure_comparison.md  
    |-- reputation_and_ratings_comparison.md
    |-- growth_5y_comparison.md
    |-- dividends_payouts_comparison.md  
    |-- china_potential_assessment.md
    `-- future_top_assets_candidates_2_3.md
\end{tcblisting}
\end{minipage}
\end{center}

\clearpage
\section{Checklists and Logs}
\label{app:checklists}
Below shows the curated checklists and an example excerpt of the log of Context Building stage. For Report Writer, there are two levels of checklists: section-wise and report-wise.

\begin{tcolorbox}[
  enhanced,
  breakable,
  colback=teal!4,
  colframe=teal!65!black,
  title={Context Builder Checklist},
  fonttitle=\bfseries
]
\begin{enumerate}
  \item \textbf{Tasks Complete}: Are all items in the \texttt{index.md} ``TODOs'' marked \texttt{[COMPLETE]}?
  \item \textbf{Hierarchy Match}: Does the directory structure of the workspace perfectly mirror the ``Target Hierarchy'' defined in \texttt{index.md}?
  \item \textbf{No Placeholders}: Are there any non-descriptive filenames (like \texttt{source\_1}, \texttt{notes.md}, etc.) existing in the knowledge base?
  \item \textbf{Full Traceability}: Does every ``Distilled Note'' contain citations (relative paths) pointing to its corresponding ``Archived Source'' file?
  \item \textbf{Exhaustive Coverage}: Can I raise a new question about the topic that cannot be fully addressed by the knowledge base? Are there any missing \textbf{region- or segment-specific information} where relevant? Are there any important aspects where you only have 1--2 weak sources?
  \item \textbf{Information Density}: Open a random \texttt{.md} file in \texttt{knowledge\_base/}. Does it contain specific data/facts, or just vague summaries? If vague, fetch again and extract details.
\end{enumerate}
\end{tcolorbox}

\begin{tcolorbox}[
  enhanced,
  breakable,
  colback=teal!4,
  colframe=teal!65!black,
  title={Report Writer: Section-level Checklist},
  fonttitle=\bfseries
]
Before you end a section-writing round, verify the following aspects with self-asking:
\begin{itemize}
  \item \textbf{Content}:
    \begin{itemize}
      \item Is the content in the \texttt{report\_outline.md} covered and the key question can be answered?
      \item Are there any parts that feel like fact listing \textbf{without} clearly telling the reader ``so what''?
    \end{itemize}
  \item \textbf{Formatting \& Style}:
    \begin{itemize}
      \item Is this section paragraph first, without abusing bullet points?
      \item Is there any data listing or comparison that can be better presented with a table?
      \item Are there tables without captions?
      \item Is the language clear, without abusing jargon and unexplained abbreviations?
      \item Is the markdown format correct? Are there unnecessary or missing indent / line breaks?
    \end{itemize}
  \item \textbf{Traceability}:
    \begin{itemize}
      \item Are there any statements or claims in the report that do not come with citations?
      \item Are citations all in correct format?
    \end{itemize}
\end{itemize}
\end{tcolorbox}

\begin{tcolorbox}[
  enhanced,
  breakable,
  colback=teal!4,
  colframe=teal!65!black,
  title={Report Writer: Report-level Checklist},
  fonttitle=\bfseries
]
After all sections are marked as \texttt{[COMPLETE]}, verify the following aspects with self-asking:
\begin{itemize}
  \item \textbf{Integrity}:
    \begin{itemize}
      \item Does \texttt{report.md} exist? Is it empty?
      \item Does \texttt{report.md} contain all three parts? (Key Takeaways, Main Body, References)
      \item Do all the sections in \texttt{report\_outline.md} appear in \texttt{report.md}?
      \item Is the markdown format correct? Are there unnecessary or missing indent / line breaks?
    \end{itemize}
  \item \textbf{Coverage}:
    \begin{itemize}
      \item List all the requirements and questions in the research task, check whether all of them are satisfied in the report.
      \item If the research task requires, is there a \textbf{time-horizon / roadmap view} that contains clear, unambiguous steps?
    \end{itemize}
  \item \textbf{Format}:
    \begin{itemize}
      \item Is the reference list placed at the end of the report as a whole?
      \item Are there duplicated citations in the reference list?
    \end{itemize}
\end{itemize}
\end{tcolorbox}

\begin{tcblisting}{
  breakable,
  colback=black!1,
  colframe=black!75,
  boxrule=0.6pt,
  arc=2pt,
  left=6pt,right=6pt,top=6pt,bottom=6pt,
  title={Log Example},
  fonttitle=\bfseries,
  listing only,
  listing options={
    basicstyle=\ttfamily\footnotesize,
    columns=fullflexible,
    keepspaces=true,
    breaklines=true
  }
}
# Round 1
Self-check (Round 1)
1) Tasks Complete? NO -- Some TODOs remain (UNECE UN R79/ALKS; NHTSA definitions; more case law).
2) Hierarchy Match? PARTIAL -- All created folders match Target Hierarchy; one planned leaf remains to be added when sources are fetched.
5) Exhaustive Coverage? NO -- Missing UNECE/NHTSA primary source coverage; limited regional case law beyond US.

Decisions for next round
- Prioritize fetching UNECE R79/ALKS official text and NHTSA/SAE official materials via alternative accessible endpoints.
- Add EU/Germany/UK ADAS-related case law where available; expand comparative matrix.

# Round 2
Workspace status summary (Round 2)
- Searched and attempted to archive UNECE (R157/ALKS) and NHTSA (ADAS definitions, SGO) sources, but target endpoints were inaccessible to the fetch tool this round.
- Left UNECE R79/ALKS and NHTSA primary materials as TODOs for next round with alternative endpoints (direct PDFs, mirrors) to be attempted.

Self-check (Round 2)
1) Tasks Complete? NO -- Outstanding TODOs: UNECE R79/ALKS extracts; NHTSA definitions/human-factors; non-US ADAS case law.
2) Hierarchy Match? PARTIAL -- International/UNECE leaf planned in index; not yet created pending source capture.
5) Exhaustive Coverage? NO -- Key international (UNECE) and US NHTSA primary references missing; regional case law beyond US still to add.

# Round 3
Workspace status summary (Round 3)
- Added international/UNECE evidence via InterRegs (ALKS R157) and ATIC (UN R79) summaries.
- Deepened comparative matrix and synthesis with UNECE and NHTSA references.

Self-check (Round 3)
1) Tasks Complete? YES -- All TODOs marked COMPLETE; comparative matrix and recommendations updated.
5) Exhaustive Coverage? YES -- Remaining primary PDFs (UNECE/NHTSA) were inaccessible to fetch, but cross-validated via reputable summaries and official SGO definitions.
[...]
\end{tcblisting}

\clearpage
\section{Benchmark Details}
\label{app:benchmark}
\paragraph{DeepResearch Bench.}
DeepResearch Bench scores a system along two complementary axes: (i) \emph{report quality} via \texttt{RACE} (Reference-based Adaptive Criteria-driven Evaluation with Dynamic weighting), and (ii) \emph{web retrieval \& citation reliability} via \texttt{FACT} (Factual Abundance and Citation Trustworthiness). In \texttt{RACE}, an LLM judge first derives task-specific weights over four orthogonal dimensions—Comprehensiveness, Insight/Depth, Instruction-Following, and Readability—by averaging weights across multiple trials, and then generates dimension-specific evaluation criteria with normalized criterion weights. The judge scores both the model report $R_{\text{tgt}}$ and a high-quality reference report $R_{\text{ref}}$ against the union of generated criteria, aggregates criterion scores into dimension scores, and forms an intermediate overall score $S_{\text{int}}(\cdot)$ by weighting dimensions with the task-specific weights. The final \texttt{RACE} score is computed as a relative score against the reference:
\[
S_{\text{final}}(R_{\text{tgt}})=\frac{S_{\text{int}}(R_{\text{tgt}})}{S_{\text{int}}(R_{\text{tgt}})+S_{\text{int}}(R_{\text{ref}})}.
\]
In \texttt{FACT}, the judge extracts statement--URL pairs from the report, deduplicates same-fact pairs per URL, retrieves the cited webpage text, and issues a binary support decision for each pair; these decisions are then summarized into citation-precision (\emph{Citation Accuracy}) and the average number of supported statement--URL pairs per task (\emph{Effective Citations}). Following the benchmark's recommended setup, we apply citation-format normalization before judging, use Gemini-2.5-Pro as the \texttt{RACE} judge and Gemini-2.5-Flash for \texttt{FACT}, and use Gemini-2.5-Pro Deep Research reports as $R_{\text{ref}}$.

\paragraph{DeepConsult.}
DeepConsult is a benchmark consisting of business and consulting prompts (e.g., market analysis, strategy, investment, and risk assessment) paired with reference reports and candidate reports in a standardized format (question, baseline answer, candidate answer). DeepConsult uses an LLM-as-a-judge \emph{pairwise} protocol: for each query, the judge compares the candidate report against a baseline reference report (commonly OpenAI Deep Research outputs) along four dimensions—Instruction Following, Comprehensiveness, Completeness, and Writing Quality—and produces a preference outcome (win/loss/tie) under randomized A/B ordering to reduce position bias. To increase robustness, the public DeepConsult evaluation described by You.com repeats judging multiple times per query (six independent trials), aggregates the outcomes into a single decision (e.g., win if preferred in at least $4/6$ trials), and reports the overall win/tie/loss rates across the benchmark (102 queries; 612 total comparisons in that setting).

\clearpage
\section{Detailed Results of DeepConsult}
\label{app:deepconsult}
Table~\ref{tab:deepconsult_detailed_gpt5} and Table~\ref{tab:deepconsult_detailed_claude} show that our system wins consistently across \textbf{Instruction Following}, \textbf{Comprehensiveness} and \textbf{Completeness} with large margins, indicating broader coverage and more thorough analyses than the baseline on most queries. The relative weakness is \textbf{Writing Quality}, suggesting that further gains may come from improving clarity, structure, and conciseness without sacrificing content coverage. Note that the winrate of \texttt{Claude} on writing quality is especially low, mainly caused by the term-heavy and citation-dense report style, which is not preferred by the judge model.

\begin{table}[h]
  \centering
  \caption{Detailed DeepConsult results of GPT-5 (pairwise comparisons against the baseline reference report). ``Net Winrate'' is computed as $\frac{\text{win}}{\text{win}+\text{lose}}$, i.e., ties are excluded.}
  \label{tab:deepconsult_detailed_gpt5}
  \footnotesize
  \begin{tabular}{lccccc}
    \toprule
    \textbf{Dimension} & \textbf{Win (\%)} & \textbf{Tie (\%)} & \textbf{Lose (\%)} & \textbf{Avg. Score} & \textbf{Net Winrate} \\
    \midrule
    Instruction Following & 79.31 & 17.24 & 3.45  & 7.33 & 0.958 \\
    Comprehensiveness     & 75.86 & 10.34 & 13.79 & 7.51 & 0.846 \\
    Completeness          & 75.86 & 24.14 & 0.00  & 8.01 & 1.000 \\
    Writing Quality       & 62.07 & 24.14 & 13.79 & 6.20 & 0.818 \\
    \midrule
    \textbf{Overall}      & \textbf{73.28} & \textbf{18.97} & \textbf{7.76} & \textbf{7.26} & \textbf{0.906} \\
    \bottomrule
  \end{tabular}
\end{table}

\begin{table}[h]
  \centering
  \caption{Detailed DeepConsult results of Claude-Sonnet-4.5 (pairwise comparisons against the baseline reference report). ``Net Winrate'' is computed as $\frac{\text{win}}{\text{win}+\text{lose}}$, i.e., ties are excluded.}
  \label{tab:deepconsult_detailed_claude}
  \footnotesize
  \begin{tabular}{lccccc}
    \toprule
    \textbf{Dimension} & \textbf{Win (\%)} & \textbf{Tie (\%)} & \textbf{Lose (\%)} & \textbf{Avg. Score} & \textbf{Net Winrate} \\
    \midrule
    Instruction Following & 100.00 & 0.00 & 0.00  & 8.56 & 1.000 \\
    Comprehensiveness     & 100.00 & 0.00 & 0.00 & 9.71 & 1.000 \\
    Completeness          & 100.00 & 0.00 & 0.00 & 9.91 & 1.000 \\
    Writing Quality       & 20.00 & 41.67 & 38.33 & 5.16 & 0.343 \\
    \midrule
    \textbf{Overall}      & \textbf{80.00} & \textbf{10.42} & \textbf{9.58} & \textbf{8.33} & \textbf{0.893} \\
    \bottomrule
  \end{tabular}
\end{table}

\clearpage
\section{Original Data for Scaling Figures}
\label{app:scaling_data}
This appendix lists the original aggregated numbers used to plot Figure~\ref{fig:scaling} in Table~\ref{tab:scaling_kb_raw} and Table~\ref{tab:scaling_drb_raw}.

\begin{table}[h]
  \centering
  \caption{Original aggregated KB statistics (averaged over 10 sampled DeepResearch Bench queries) used in Figure~\ref{fig:scaling} (left). ``Citation count'' is computed as the maximum citation index \texttt{[x]} appearing in \texttt{report.md}.}
  \label{tab:scaling_kb_raw}
  \footnotesize
  \begin{tabular}{lccccc}
    \toprule
    \textbf{Rounds} & \textbf{\#sources} & \textbf{\#non-sources} & \textbf{\#unique URLs} & \textbf{report chars} & \textbf{citation count} \\
    \midrule
    3  & 21.6 & 17.1 & 21.3 & 21306.2 & 18.5 \\
    5  & 33.3 & 18.0 & 32.1 & 25094.5 & 26.8 \\
    10 & 38.8 & 20.5 & 38.0 & 29244.5 & 37.1 \\
    \bottomrule
  \end{tabular}
\end{table}

\begin{table}[h]
  \centering
  \caption{Original averaged performance scores used in Figure~\ref{fig:scaling} (right).}
  \label{tab:scaling_drb_raw}
  \footnotesize
  \begin{tabular}{lccccc}
    \toprule
    \textbf{Rounds} & \textbf{Comp.} & \textbf{Insight} & \textbf{Instr.} & \textbf{Readability} & \textbf{RACE} \\
    \midrule
    3  & 49.72 & 52.27 & 51.41 & 51.43 & 51.18 \\
    5  & 51.96 & 53.43 & 51.73 & 51.93 & 52.37 \\
    10 & 52.31 & 54.70 & 52.67 & 51.66 & 53.05 \\
    \bottomrule
  \end{tabular}
\end{table}

\section{Cost}
\label{app:cost}
We report the average cost of FS-Researcher under different models and context-building budgets (3/5/10 rounds) on the 10 sampled DeepResearch Bench queries used in Section~\ref{sec:context-builder}. Table~\ref{tab:cost} shows the statistics. As expected, allocating more rounds increases the LLM cost (6.10 $\rightarrow$ 8.16 $\rightarrow$ 12.54 \$/query), reflecting more test-time computation spent in the Context Builder. The higher cost of \texttt{Claude-Sonnet-4.5} is due to higher token price.
The web-browsing tool calls also grow with the budget. Notably, \texttt{search\_web} grows faster than \texttt{read\_webpage} when moving from 5 to 10 rounds (+72\% vs.\ +28\%), and the read-to-search ratio decreases from 1.18 (5 rounds) to 0.88 (10 rounds). This suggests that extra budget is increasingly used to expand the breadth of evidence acquisition (more query reformulations and gap-driven searches), while page reading saturates with more rounds.
This scaling of tool calls also provides an intuitive source of test-time scaling: higher budgets induce more evidence collection, which supports the quality gains in Figure~\ref{fig:scaling} (right).

\begin{table}[h]
  \centering
  \caption{Cost and web-browsing tool usage of FS-Researcher under different context-building budgets. Values are averaged over 10 sampled DeepResearch Bench queries.}
  \label{tab:cost}
  \footnotesize
  \begin{tabular}{lcccc}
    \toprule
    \multirow{2}{*}{\textbf{Budget}} & \textbf{LLM Cost} & \multirow{2}{*}{\textbf{\#search\_web}} & \multirow{2}{*}{\textbf{\#read\_webpage}} & \multirow{2}{*}{\textbf{\#sources}} \\
    & \textbf{(\$/query)} &  &  &  \\
    \midrule
    10 rounds (GPT) & 12.54 & 66.25 & 58.00 & 38.20 \\
    5 rounds (GPT)  & 8.16 & 38.42 & 45.33 & 33.30 \\
    3 rounds (GPT)  & 6.10  & 28.62 & 32.42 & 21.60 \\
    3 rounds (Claude) & 9.31 & 36.08 & 38.20 & 24.73 \\
    \bottomrule
  \end{tabular}
\end{table}

\clearpage
\section{Tool Usage Patterns}
\label{app:tool_usage_patterns}
We analyze the frequency--position relationship of tool calls in the first three iterations of both stages (Context Building and Report Writing), where the x-axis is the normalized position within a trajectory and the y-axis is the tool type (color indicates row-normalized frequency).
We observe the following consistent patterns:
\begin{itemize}
  \item \textbf{\texttt{read\_webpage} lags behind \texttt{search\_web}.} In Context Building, \texttt{search\_web} tends to appear earlier than \texttt{read\_webpage}, matching the natural ``search-then-read'' workflow.
  \item \textbf{\texttt{ls} concentrates at the beginning (and sometimes the end).} In both stages, \texttt{ls} is highly concentrated near the start of an iteration, suggesting the agent first confirms the workspace state; a smaller amount of \texttt{ls} also appears near the end, consistent with checklist-driven self-inspection.
  \item \textbf{Early reading differs by stage and iteration.} In Context Building, iteration~1 shows \texttt{read} occurring more frequently in the latter half, while iterations~2--3 shift \texttt{read} toward the early part of the trajectory, indicating that later iterations begin with evaluating the existing workspace before further browsing/writing. In Report Writing, \texttt{read} concentrates in the first half across iterations because this stage does not use web tools and relies on the knowledge base as the sole information source.
  \item \textbf{Late edits: \texttt{replace} and \texttt{delete} cluster toward the end.} Across both stages, \texttt{replace} and \texttt{delete} are more prevalent later in the trajectory, reflecting post-hoc self-check and error-fixing behaviors.
\end{itemize}

\begin{figure*}[!htbp]
  \centering
  \includegraphics[width=\linewidth]{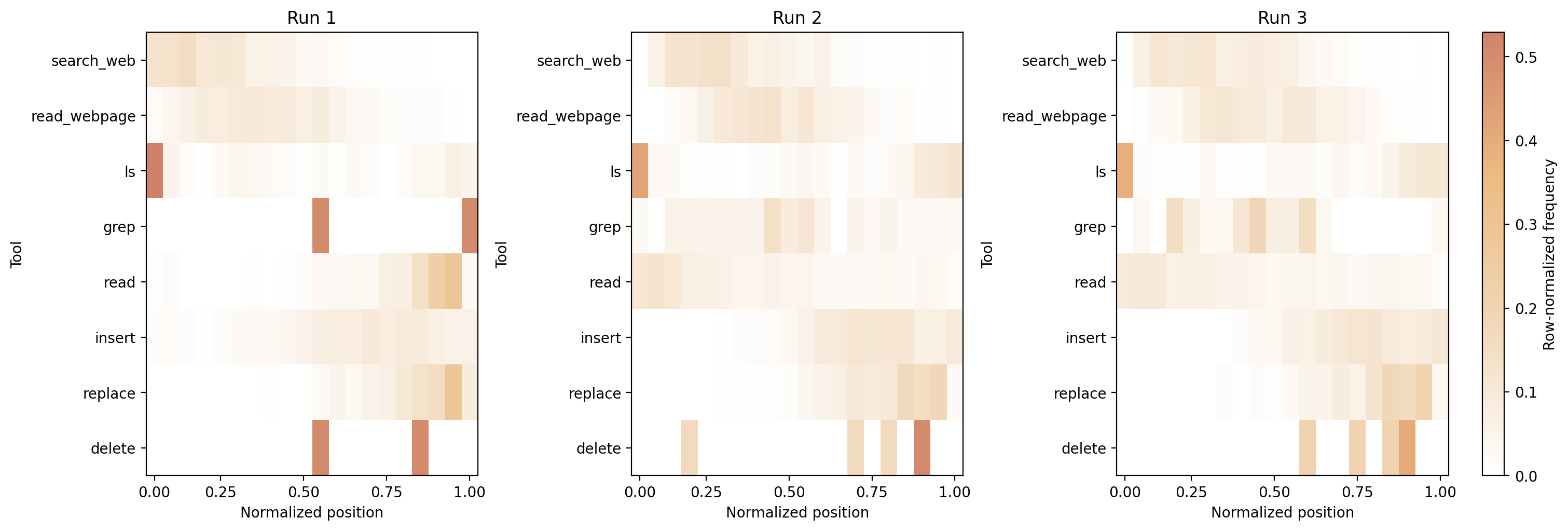}
  \caption{Tool usage heatmap for the Context Building stage (first three iterations).}
  \label{fig:tool_usage_heatmap_context_building}
\end{figure*}

\begin{figure*}[!htbp]
  \centering
  \includegraphics[width=\linewidth]{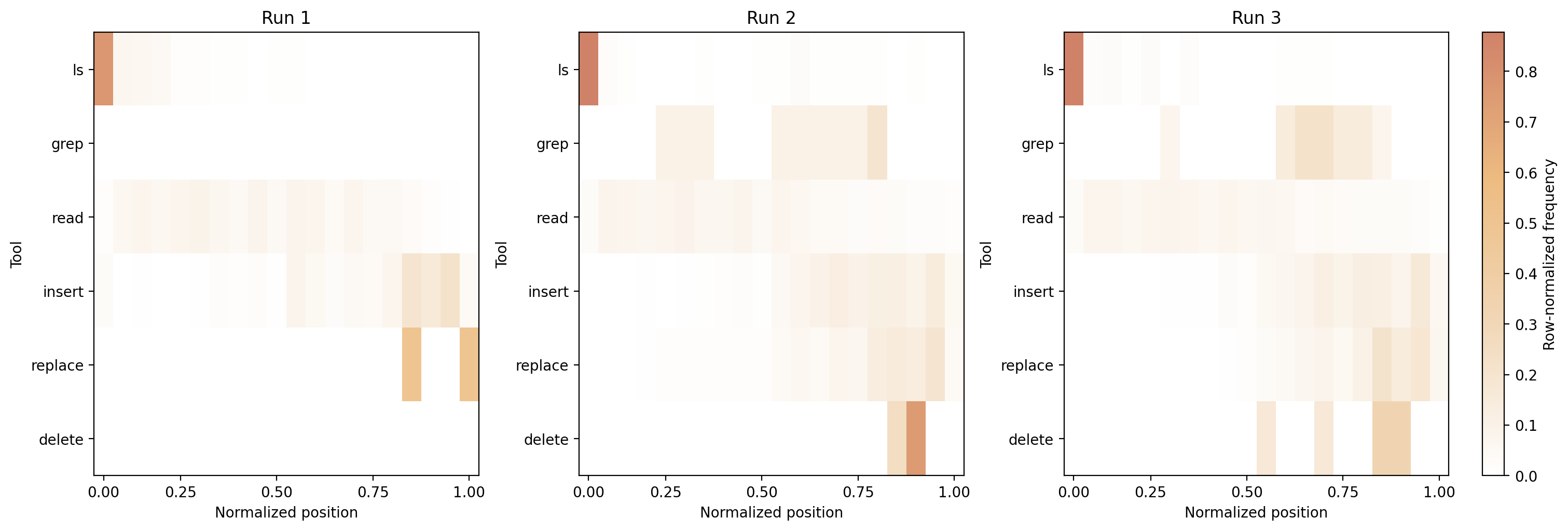}
  \caption{Tool usage heatmap for the Report Writing stage (first three iterations).}
  \label{fig:tool_usage_heatmap_report_writing}
\end{figure*}

\clearpage
\section{Case Study (3/5/10-round KB Growth and Its Impact on Reports)}
\label{app:case_study}

This appendix compares the research artifacts produced for the same query under three context-building budgets (3/5/10 rounds).
We show (i) simplified KB directory structures (with minor cosmetic cleanup such as omitting redundant filenames) and (ii) short report excerpts.
When an excerpt is originally written in Chinese in the artifact, we provide an English translation for readability.

\begin{tcblisting}{
  breakable,
  colback=black!1,
  colframe=black!75,
  boxrule=0.6pt,
  arc=2pt,
  left=6pt,right=6pt,top=6pt,bottom=6pt,
  title={KB Structure (3 rounds): 47 distilled notes; 24 archived sources},
  fonttitle=\bfseries,
  listing only,
  listing options={
    basicstyle=\ttfamily\footnotesize,
    columns=fullflexible,
    keepspaces=true,
    breaklines=true
  }
}
./knowledge_base/
|-- sources/
|-- global_insurers/
|   |-- rankings/
|   `-- comparative_metrics/ 
`-- company_dossiers/
\end{tcblisting}

\begin{tcblisting}{
  breakable,
  colback=black!1,
  colframe=black!75,
  boxrule=0.6pt,
  arc=2pt,
  left=6pt,right=6pt,top=6pt,bottom=6pt,
  title={KB Structure (5 rounds): 75 distilled notes; 55 archived sources},
  fonttitle=\bfseries,
  listing only,
  listing options={
    basicstyle=\ttfamily\footnotesize,
    columns=fullflexible,
    keepspaces=true,
    breaklines=true
  }
}
./knowledge_base/
|-- sources/
|-- global_rankings/
`-- dimensions/
    |-- financing_and_solvency/
    |-- reputation/
    |-- dividends/
    |-- growth_5y/
    `-- china_market_potential/ 
\end{tcblisting}

\begin{tcblisting}{
  breakable,
  colback=black!1,
  colframe=black!75,
  boxrule=0.6pt,
  arc=2pt,
  left=6pt,right=6pt,top=6pt,bottom=6pt,
  title={KB Structure (10 rounds): 98 distilled notes; 59 archived sources},
  fonttitle=\bfseries,
  listing only,
  listing options={
    basicstyle=\ttfamily\footnotesize,
    columns=fullflexible,
    keepspaces=true,
    breaklines=true
  }
}
./knowledge_base/
|-- sources/
|-- global_insurance_landscape/
|   |-- top10_rankings/
|   `-- metrics_definitions/
|-- company_profiles/
|   |-- overview_and_financials_5y
|   |-- dividends_and_payouts
|   |-- credit_ratings
|   `-- China presence
`-- comparative_analysis/ 
\end{tcblisting}

\newpage
\begin{tcolorbox}[
  enhanced,
  breakable,
  colback=teal!4,
  colframe=teal!65!black,
  title={Report Excerpts},
  fonttitle=\bfseries
]
\small
\textbf{3 rounds}

``Current landscape: by assets, the top ten are Allianz, Berkshire Hathaway, China Life, Ping An, Prudential Financial (US), AXA, MetLife, Legal \& General, Manulife, and Nippon Life.''\par
\medskip
\textbf{5 rounds}

``The 2--3 companies most likely to remain among the global asset leaders are Allianz, China Life, and Ping An Group. The justification includes ranks by `net non-banking assets' (Allianz \#1, China Life \#3, Ping An \#4) and capital strength (e.g., Allianz group Solvency II 209\%) alongside a progressive dividend policy and buybacks.''\par
\medskip
\textbf{10 rounds}

``Capital and financing (`safety cushion'): many leaders maintain substantial capital coverage at the end of the period, supporting counter-cyclical investment and stable shareholder returns, e.g., Allianz Solvency II 209\%, AXA Solvency II 216\%, Zurich SST $\sim$218\%, AIA shareholder capital ratio 236\%, Ping An / China Life comprehensive solvency $\sim$204\% / 208\%, Prudential plc GWS-to-GPCR coverage $\sim$280\%, and Manulife (MLI) LICAT $\sim$137\%.''\par
``Final candidates (2--3 most likely to rank high in future total assets and overall strength): AIA, Allianz, and Manulife.''\par
\end{tcolorbox}

\clearpage
\section{Agent Harness Contribution}
\label{app:harness_contribution}
To disentangle the contribution of the agent harness from the backbone model, we compare Gemini-2.5-Pro under three configurations: (1) the bare model with only a search tool, (2) the model inside its official deep-research harness, and (3) the model inside FS-Researcher. As shown in Table~\ref{tab:harness_contribution}, the official harness already provides a dramatic improvement over the bare model (+17.81 RACE), confirming that an agent harness is essential for deep research. FS-Researcher further improves upon the official harness (+2.80 RACE), with particularly large gains on \texttt{Insight} (+5.58), demonstrating that our framework provides significant additional value on top of the foundational model's capabilities.

\begin{table}[h]
  \centering
  \caption{DeepResearch Bench performance of Gemini-2.5-Pro under different configurations, showing the contribution of the agent harness.}
  \resizebox{0.75\columnwidth}{!}{
  \begin{tabular}{lccccc}
    \toprule
    \textbf{Gemini-2.5-Pro} & \textbf{Comp.} & \textbf{Insight} & \textbf{Instr.} & \textbf{Read.} & \textbf{RACE} \\
    \midrule
    with search tool only   & 31.75 & 24.61 & 40.24 & 32.76 & 31.90 \\
    with official harness   & 49.51 & 49.45 & 50.12 & 50.00 & 49.71 \\
    with FS-Researcher      & \textbf{51.25} & \textbf{55.03} & \textbf{52.38} & \textbf{50.77} & \textbf{52.51} \\
    \bottomrule
  \end{tabular}
  }
  \label{tab:harness_contribution}
\end{table}

\section{Accessibility to Smaller Models}
\label{app:smaller_models}
To assess whether FS-Researcher is usable beyond top-tier LLMs, we run the framework with \texttt{GPT-5-mini} on the 10 sampled DeepResearch Bench queries used in Section~4. We allow 10 context-building rounds to compensate for the model's weaker agentic capabilities. As shown in Table~\ref{tab:smaller_models}, GPT-5-mini achieves performance comparable to OpenAI-DeepResearch (46.63 vs.\ 46.45 RACE) at significantly lower cost (\$2.51 vs.\ \$6.10/query with GPT-5). While a gap remains compared to GPT-5 under the same framework, this result demonstrates that the framework is accessible to smaller, more cost-efficient models.

\begin{table}[h]
  \centering
  \caption{DeepResearch Bench performance of GPT-5-mini under FS-Researcher compared with OpenAI-DeepResearch and GPT-5. Cost is the average LLM API cost per query.}
  \resizebox{0.75\columnwidth}{!}{
  \begin{tabular}{lcccccc}
    \toprule
    \textbf{Method} & \textbf{Comp.} & \textbf{Insight} & \textbf{Instr.} & \textbf{Read.} & \textbf{RACE} & \textbf{Cost} \\
    \midrule
    OpenAI-DeepResearch               & 46.46 & 43.73 & 49.39 & 47.22 & 46.45 & - \\
    FS-Researcher (GPT-5, 3 rounds)   & 49.72 & 52.27 & 51.41 & 51.43 & 51.18 & \$6.10 \\
    FS-Researcher (GPT-5-mini, 10 rounds) & 44.42 & 46.89 & 48.50 & 47.59 & 46.63 & \$2.51 \\
    \bottomrule
  \end{tabular}
  }
  \label{tab:smaller_models}
\end{table}

\section{Readability Recovery via Post-hoc Rephrasing}
\label{app:readability_rephrasing}
As discussed in Section~4.1, the readability drop at 10 rounds is a presentation-level artifact caused by the Report Writer adopting a denser, more technical style when the KB is large. To verify that this is recoverable, we add a post-processing rephrasing pass in the Report Writing stage on the 10-round setting. The rephrasing instruction is:

\begin{quote}
\small
\emph{Rephrase the given report based on the knowledge base. Specifically: (1) Increase narrative context, illustrative examples, and scenario analysis to enhance accessibility for broader audiences. (2) Reduce jargon and define technical terms to improve readability for non-experts. (3) Ensure all sub-tasks in the query are explicitly addressed, avoiding over-compression of information. (4) Balance concise, data-driven summaries with richer explanations and transitions for better engagement and comprehension. \textbf{ATTENTION}: Mainly improve report style and organization. Do not modify the core ideas and arguments.}
\end{quote}

As shown in Table~\ref{tab:readability_rephrasing}, the rephrasing pass recovers readability (51.66 $\rightarrow$ 51.92) with negligible impact on other dimensions, confirming that comprehensiveness and readability are not inherently in conflict. We do not incorporate this step into our default framework, as the specific rephrasing instructions are derived from benchmark-specific observations rather than a general-purpose mechanism.

\begin{table}[h]
  \centering
  \caption{Effect of post-hoc rephrasing on the 10-round setting. Readability is recovered while other dimensions remain stable.}
  \resizebox{0.75\columnwidth}{!}{
  \begin{tabular}{lccccc}
    \toprule
    \textbf{Setting} & \textbf{Comp.} & \textbf{Insight} & \textbf{Instr.} & \textbf{Read.} & \textbf{RACE} \\
    \midrule
    10 rounds                 & 52.31 & 54.70 & 52.67 & 51.66 & 53.05 \\
    10 rounds (w/ rephrasing) & 52.25 & 54.88 & 52.60 & \textbf{51.92} & \textbf{53.17} \\
    \bottomrule
  \end{tabular}
  }
  \label{tab:readability_rephrasing}
\end{table}

\section{Cost Reduction via Context Compression}
\label{app:context_compression}
Our framework is orthogonal to and readily combinable with existing context compression methods. We implement a simple compression strategy: raw webpages are summarized by a smaller model (\texttt{GPT-5-mini}) and only the summaries are retained in the main agent's context. Table~\ref{tab:context_compression} shows that this reduces Context Builder cost by 47\% (\$3.77 $\rightarrow$ \$2.00) with negligible performance drop (51.18 $\rightarrow$ 51.14 RACE). Moreover, with recent models such as Kimi-K2.5 and Minimax-M2.5 offering API costs at 1/5 or less of GPT-5 while exhibiting comparable planning and search capabilities, it is increasingly practical to deploy FS-Researcher at substantially lower cost.

\begin{table}[h]
  \centering
  \caption{Effect of context compression using GPT-5-mini as summarizer. ``CB Cost'' denotes the average Context Builder LLM cost per query.}
  \resizebox{0.75\columnwidth}{!}{
  \begin{tabular}{lcccccc}
    \toprule
    \textbf{Setting} & \textbf{Comp.} & \textbf{Insight} & \textbf{Instr.} & \textbf{Read.} & \textbf{RACE} & \textbf{CB Cost} \\
    \midrule
    3 rounds                        & 49.72 & 52.27 & 51.41 & 51.43 & 51.18 & \$3.77 \\
    3 rounds (w/ compression) & 49.37 & 52.13 & 52.05 & 51.44 & 51.14 & \$2.00 \\
    \bottomrule
  \end{tabular}
  }
  \label{tab:context_compression}
\end{table}

\section{Wall-clock Latency Profiling}
\label{app:latency_profiling}
A potential concern with file-system-based agents is the latency introduced by frequent file I/O operations. We profile the wall-clock time on the 10 sampled queries from Section~4. As shown in Table~\ref{tab:latency}, file I/O accounts for less than 0.03\% of total time in both stages, while LLM inference and web browsing dominate. This confirms that the file-system workspace introduces negligible latency overhead---a worthwhile trade-off for the persistence and cross-session scalability validated by our scaling experiments (Section~4.1).

\begin{table}[h]
  \centering
  \caption{Average wall-clock time breakdown per query. File I/O accounts for $<$0.03\% of total time; latency is dominated by LLM inference and web browsing.}
  \small
  \begin{tabular}{lccc}
    \toprule
    \textbf{Stage} & \textbf{LLM Calls} & \textbf{Web Tools} & \textbf{File I/O} \\
    \midrule
    Context Building & 236.72s & 219.96s & 0.055s \\
    Report Writing   & 376.65s & 0s      & 0.186s \\
    \bottomrule
  \end{tabular}
  \label{tab:latency}
\end{table}

\end{document}